\documentclass[sigconf]{acmart}

\AtBeginDocument{%
  }
\usepackage{algorithm}
\usepackage{algorithmic}
\usepackage{enumitem}
\usepackage{subfigure}  % 替代 subfigure
\usepackage{arydshln}
\usepackage{pgfplots}
\pgfplotsset{width=8cm,compat=1.13}
\usepackage{csvsimple}
\usepackage{tikz}
\usepackage{enumitem} % 导言区
\usepackage{array}   % 必须：支持 >{\centering}
\usepackage{adjustbox}
\definecolor{mycolorblack}{RGB}{84,89,105}
\newcommand{\name}{MDTI}

\copyrightyear{2026}
\acmYear{2026}
\setcopyright{cc}
\setcctype{by}
\acmConference[WWW '26]{Proceedings of the ACM Web Conference 2026}{April 13--17, 2026}{Dubai, United Arab Emirates}
\acmBooktitle{Proceedings of the ACM Web Conference 2026 (WWW '26), April 13--17, 2026, Dubai, United Arab Emirates}
\acmPrice{}
\acmDOI{10.1145/3774904.3792388}
\acmISBN{979-8-4007-2307-0/2026/04}
\begin{document}

\title{Multimodal Trajectory Representation Learning for Travel Time
Estimation}


\author{Zhi Liu}
\orcid{0000-0001-8320-820X}
\affiliation{%
  \institution{Zhejiang University of Technology}
 \department{College of Computer Science and Technology, Zhejiang Key Laboratory of Visual Information Intelligent Processing}
  \city{Hangzhou}
  \postcode{310023}
  \country{China}}
\email{lzhi@zjut.edu.cn}

\author{Xuyuan Hu}
\orcid{0009-0008-5678-2124}
\affiliation{%
  \institution{Zhejiang University of Technology}
  \department{College of Computer Science and Technology}
  \city{Hangzhou}
  \postcode{310023}
  \country{China}}
\email{huxvyuan@gmail.com}

\author{Xiao Han}
\authornote{Corresponding author.}
\orcid{0000-0002-3478-964X}
\affiliation{%
  \institution{Zhejiang University of Technology}
  \department{College of Computer Science and Technology, Zhejiang Key Laboratory of Visual Information Intelligent Processing}
  \city{Hangzhou}
  \postcode{310023}
  \country{China}}
\email{hahahenha@gmail.com}

\author{Zhehao Dai}
\orcid{0009-0002-2685-0161}
\affiliation{%
  \institution{Zhejiang University of Technology}
  \department{College of Computer Science and Technology}
  \city{Hangzhou}
  \postcode{310023}
  \country{China}}
\email{zhehaodai@outlook.com}

\author{Zhaolin Deng}
\orcid{0009-0000-5968-7888}
\affiliation{%
  \institution{Zhejiang University of Technology}
  \department{College of Computer Science and Technology}
  \city{Hangzhou}
  \postcode{310023}
  \country{China}}
\email{211123120032@zjut.edu.cn}

\author{Guojiang Shen}
\orcid{0000-0003-1064-1250}
\affiliation{%
  \institution{Zhejiang University of Technology}
  \department{College of Computer Science and Technology, Zhejiang Key Laboratory of Visual Information Intelligent Processing}
  \city{Hangzhou}
  \postcode{310023}
  \country{China}
}
\email{gjshen1975@zjut.edu.cn}

\author{Xiangjie Kong}
\orcid{0000-0003-2698-3319}
\affiliation{%
  \institution{Zhejiang University of Technology}
  \department{College of Computer Science and Technology, Zhejiang Key Laboratory of Visual Information Intelligent Processing}
  \city{Hangzhou}
  \postcode{310023}
  \country{China}
}
\email{xjkong@ieee.org}

\renewcommand{\shortauthors}{Zhi Liu et al.}

\begin{abstract}
Accurate travel time estimation (TTE) plays a crucial role in intelligent transportation systems. However, it remains challenging due to heterogeneous data sources and complex traffic dynamics. 
% Moreover, conventional approaches typically convert trajectory data into fixed-length representations, neglecting the inherent variability of real-world motion patterns, which often leads to information loss or feature redundancy. 
Moreover, traditional approaches typically convert trajectory data into fixed-length representations. This overlooks the inherent variability of real-world motion patterns, often resulting in information loss and redundancy.
To address these challenges, this paper introduces the Multimodal Dynamic Trajectory Integration (MDTI) framework--a novel multimodal trajectory representation learning approach that integrates GPS sequences, grid trajectories, and road network constraints to enhance the performance of TTE. MDTI employs modality-specific encoders and a multimodal fusion module to capture complementary spatial, temporal, and topological semantics, while a dynamic trajectory modeling mechanism adaptively regulates information density for trajectories of varying lengths. Two self-supervised pretraining objectives, named contrastive alignment and masked language modeling, further strengthen multimodal consistency and contextual understanding. Extensive experiments on three real-world datasets demonstrate that MDTI consistently outperforms state-of-the-art baselines, confirming its robustness and strong generalization abilities. The code is publicly available at: \url{https://github.com/City-Computing/MDTI}.

% \url{https://github.com/freshhxy/MDTI/}
% \url{https://anonymous.4open.science/r/MDTI-3433/}
% To address these issues, this paper proposes a Multimodal Dynamic Trajectory Integration model (MDTI), which achieves comprehensive trajectory representation by jointly modeling GPS sequences, grid trajectories, and road network constraints. We design a cross-modal interaction module to facilitate semantic alignment and information complementarity across modalities, and introduce a dynamic trajectory extraction mechanism to adaptively encode features of trajectories with varying lengths, avoiding biases inherent in fixed-length representations. Extensive experiments on multiple real-world datasets demonstrate that the proposed model effectively improves the accuracy of travel time estimation and outperforms existing representative methods.\\

% \noindent\textbf{Relevance Statement:} 
% % This paper focuses on integrating GPS sequences, grid trajectories, and road network semantic information to dynamically construct multimodal trajectory representations for travel time estimation. The proposed method addresses the challenges of semantic fusion and dynamic representation in large-scale web-based mobility data and can assist with semantics and knowledge.
% This paper proposes a method to integrate open-source GPS sensor data, grid trajectories, and road network semantics for dynamic multimodal trajectory representation learning, addressing semantic fusion and representation challenges in large-scale mobility data on the web.

\end{abstract}



\maketitle

\section{Introduction}

% \textcolor{blue}{With the rapid development of smart cities and the widespread adoption of travel service applications~\cite{Elassy_Al-Hattab_Takruri_Badawi_2024}, an unprecedented volume of trajectory data has been generated in cyberspace.These massive, fine-grained datasets, continuously produced by vehicles, mobile devices, and various smart terminals, provide a valuable foundation for gaining deeper insights into urban dynamics, optimizing transportation systems, and building intelligent travel services. Against this backdrop,} 
% With the development of connected vehicle technology, massive amounts of trajectory-based sensor data are generated and exposed on the Web.
% In this context, travel time estimation (TTE)~\cite{han2023ieta}, as a key task in traffic flow analysis, can calculate these data into a unified indicator and directly participate in the decision-making tasks of traffic route planning, induction, and control.
With the rapid development of connected vehicle technologies, vast amounts of trajectory-based sensor data are continuously generated and shared on the Web. Within this context, travel time estimation (TTE)~\cite{han2023ieta} has emerged as a core task in traffic flow analysis~\cite{zhang2017deep, zheng2015trajectory,liu2024traffic,han2024deep}. By transforming heterogeneous trajectory data into a unified indicator, TTE provides essential input for downstream decision-making processes such as route planning, traffic guidance, and can also support signal control~\cite{han2023mitigating}.
% travel time estimation~\cite{han2023ieta}has become a vital element of traffic route planning, guidance, and control systems, attracting increasing attention from both researchers and urban planners.
% Travel time estimation (TTE) 
% TTE involves predicting the time required to travel a specific route from a given starting point to a destination, while accounting for various dynamic factors, such as traffic conditions, road characteristics, and environmental influences~\cite{liu2023uncertainty}. 
% Accurate travel time predictions offer significant benefits, including optimized route selection, reduced congestion, and improved resource management in urban areas.
% Key aspects of TTE include its reliance on real-time data, multimodal traffic analysis, and its ability to adapt to fluctuating traffic patterns.
% However, the primary difficulty of TTE lies in its difficulty in addressing the complexity of multimodal spatiotemporal traffic systems.
% This highlights the need for advanced methodologies to enhance the precision and reliability of TTE.
TTE focuses on predicting the time required to traverse a specific route from an origin to a destination, while accounting for dynamic factors such as traffic conditions~\cite{shen2021attention}, road characteristics, and environmental influences~\cite{liu2023uncertainty}. Accurate estimation enables more efficient route planning~\cite{chen2025sustainability}, congestion mitigation~\cite{han2020congestion}, and signal control~\cite{han2023mitigating} in urban transportation systems. Core aspects of TTE include its dependence on real-time data, integration of multimodal traffic information, and adaptability to rapidly changing traffic patterns. The main challenge, however, lies in effectively modeling the inherent complexity of multimodal spatiotemporal traffic systems. This underscores the need for advanced approaches to improve the precision and robustness of TTE.

\begin{figure}
\subfigure[Multi-modal trajectory]{
\includegraphics[width=0.47\linewidth]{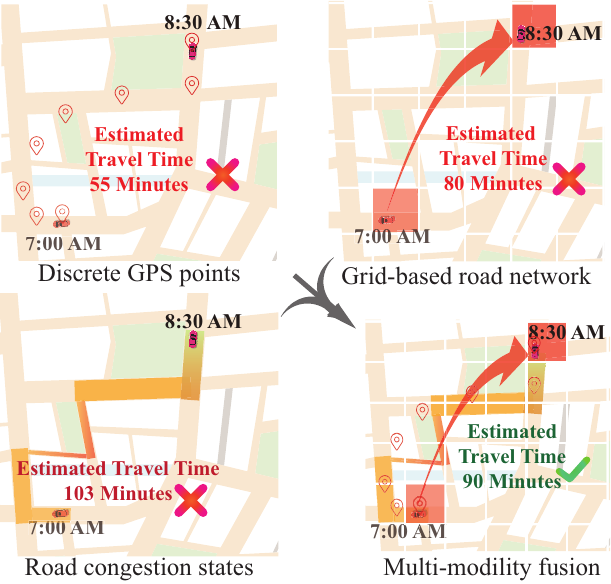}
\label{fig:1a}
}
\subfigure[Dynamic trajectory length]{
\includegraphics[width=0.47\linewidth]{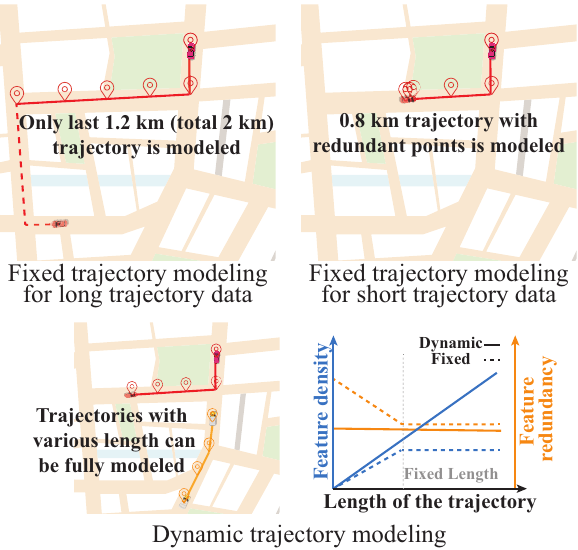}
\label{fig:1b}
}
\caption{Toy examples of online trajectory data.}
\label{fig:1}
\end{figure}

Trajectory representation learning~\cite{li2018deep,yao2017trajectory} provides an effective framework for modeling multimodal spatiotemporal data, playing a crucial role in reconstructing complex trajectory patterns and supporting TTE tasks. Early trajectory representation learning (TRL) approaches mainly employed univariate sequence-to-sequence architectures~\cite{hochreiter1997long,vaswani2017attention} to encode raw GPS trajectories~\cite{fu2020trembr,han2025garlic}. In these methods, trajectories are modeled as sequences of time-stamped data points, where each point contains a spatial coordinate pair (latitude and longitude), optionally accompanied by other attributes such as speed. Although such models effectively capture temporal dependencies and map sequential data into latent embeddings, they fall short in representing multimodal trajectory information, thereby limiting their applicability and robustness in real-world traffic scenarios.

As illustrated in Figure~\ref{fig:1a}, discrete GPS trajectory data often suffers from drift (top left), grid-based road network modeling leads to limited accuracy (top right), and congestion-based analysis depends heavily on real-time updates (bottom left). Consequently, relying on a single-source trajectory data representation is insufficient for accurate and reliable travel time estimation. In contrast, integrating multiple information sources within a unified framework enables more precise and robust prediction of vehicle travel times, as demonstrated in the bottom right of Figure~\ref{fig:1a}. Therefore, extracting semantic structures from different modalities and achieving effective alignment and fusion is a major challenge.

Another challenge in trajectory representation lies in converting dynamically varying trajectory samples into fixed-length feature vectors. For instance, GPS data collected from different vehicles often differ in spatiotemporal sampling frequencies~\cite{andersen2017sampling}, sequence lengths~\cite{wang2018complete}, and other factors. Many existing methods address trajectory variability by setting a maximum sequence length~\cite{jiang2023self}, padding shorter trajectories with zeros, and clipping longer ones for representation learning. However, these strategies often introduce redundant noise~\cite{chen2024deep} or cause information loss. As shown in Figure~\ref{fig:1b}, when trajectories exceed the maximum threshold (top left), only the last segment is retained, preventing a full representation of the overall trajectory. Conversely, shorter trajectories (top right) are padded with redundant GPS points, which overemphasize local features such as the starting point and hinder effective representation learning. To overcome these issues, dynamic modeling techniques (bottom) can adaptively generate embeddings with differentiated information density and reduced redundancy as trajectory lengths~\cite{berndt1994using, chen2004marriage, chen2005robust} vary. Therefore, The second challenge is encoding trajectories of varying lengths and structures into uniform fixed-length feature representations.

To address these challenges, we propose the Multimodal Dynamic Trajectory Integration Model (\name), which jointly models GPS sequences, grid trajectories, and road network constraints to comprehensively represent trajectory features for travel time estimation. The proposed framework incorporates a multimodal fusion module that captures relationships among different modalities, enabling effective alignment and consistent semantic extraction. In addition, dedicated modality-specific encoders map heterogeneous inputs into a unified feature space, facilitating seamless multimodal fusion. Furthermore, a dynamic trajectory extraction module adaptively adjusts the information density of trajectory feature vectors according to trajectory length, preventing feature over-concentration in local embeddings and promoting balanced and discriminative representations. 
The main contributions of this work are summarized as follows:

\begin{itemize}[itemsep=0pt,parsep=0pt,topsep=0pt,leftmargin=*]
    \item We propose the MDTI framework, which jointly leverages GPS sequences, grid trajectories, and road network constraints to systematically overcome the limitations of existing methods that rely on single or dual modalities, thereby enhancing semantic richness and completeness in trajectory representation.
    
    \item We designed a multimodal fusion module that not only aligns heterogeneous features and extracts consistent cross-modal semantics but also adaptively encodes trajectories of varying lengths, mitigating the inherent information loss and redundancy issues in fixed-length modeling.
    
    \item Extensive experiments on multiple real-world transportation datasets demonstrate that \name\  significantly outperforms state-of-the-art methods in TTE, validating the effectiveness of multimodal fusion and dynamic modeling in addressing the core challenges of trajectory representation.
\end{itemize}
% \section{Overview}

\section{Preliminary}

\textbf{Definition 1 (GPS Trajectory).}  
A GPS trajectory is defined as an ordered sequence of raw GPS points collected at a fixed sampling interval by a GPS-enabled device:
\begin{equation}
\mathcal{T}^{gps} = \{p_i\}_{i=1}^T, \quad p_i = (x_i, y_i, t_i),
\end{equation}
where $x_i$ and $y_i$ denote longitude and latitude, and $t_i$ denotes the timestamp. Here, $T$ is the total number of points in the trajectory.
\textbf{Definition 2 (Grid Trajectory).}  
A grid trajectory $\mathcal{T}^{grid}$ maps GPS points onto a discretized spatial grid. 
Let the region of interest be partitioned into $M \times N$ cells, each representing a spatial unit. 
The trajectory is then expressed as a sequence of grid cell identifiers:
\begin{equation}
\mathcal{T}^{grid} = \{ g_t \}_{t=1}^{T}, \quad g_t \in \{1, \ldots, M \times N\},
\end{equation}
where $g_t$ denotes the identifier (ID) of the grid cell visited at time $t$. 
\textbf{Definition 3 (Road Network).}  
A road network is represented as a directed graph $\mathcal{G} = (\mathcal{V}, \mathcal{E})$,  
where $\mathcal{V} = \{ v_1, v_2, \ldots, v_{|\mathcal{V}|} \}$ denotes the set of vertices, and each vertex $v_i \in \mathcal{V}$ corresponds to a distinct road segment. $\mathcal{E} \subseteq \mathcal{V} \times \mathcal{V}$ denotes the set of directed edges, where each edge $e_{ij} = (v_i, v_j) \in \mathcal{E}$ represents the navigable connection from road segment $v_i$ to $v_j$.\\
\textbf{Definition 4 (Road Trajectory).}  
A road trajectory $\mathcal{T}^{road}$ is a timestamped sequence of road segments obtained by map-matching GPS points $\mathcal{T}^{gps}$ to the road network $\mathcal{G}$. 
It can be expressed as
\begin{equation}
\mathcal{T}^{road} = \{ v_t \}_{t=1}^{T}, \quad v_t \in \mathcal{V},
\end{equation}
where $v_t$ denotes the road segment visited at time $t$.

\section{Methodology}

% \hx{
% In this section, we present a detailed and comprehensive description of the \name\ framework.
% The overall architecture of \name\ is illustrated in Figure~\ref{fig:model}.
% Three different encoders (i.e., Grid Encoder, GPS Encoder, and Road Encoder) are designed to process trajectory data of different modalities, giving the model preliminary multimodal processing capabilities.
% Among these, the graph modeling and analysis capabilities of GAT and GNN are leveraged to encode trajectories based on map matching, while the robust semantic analysis capabilities of the large model further enhance the representation of raw GPS data.
% Additionally, a multimodal fusion module is implemented to align and integrate trajectory information, which is then used to estimate vehicle travel time.
% Furthermore, an innovative dynamic trajectory alignment method is employed to enrich the information content of feature embeddings in trajectory representation learning, thereby improving the accuracy of downstream travel time estimation (TTE) tasks.
% Finally, two different loss functions are deployed to ensure the stability of model training.
% The detailed introduction of modules in the proposed \name\ are illustrated in the following subsections.
% }

\begin{figure*}[t]
  \centering
  \includegraphics[width=\textwidth]{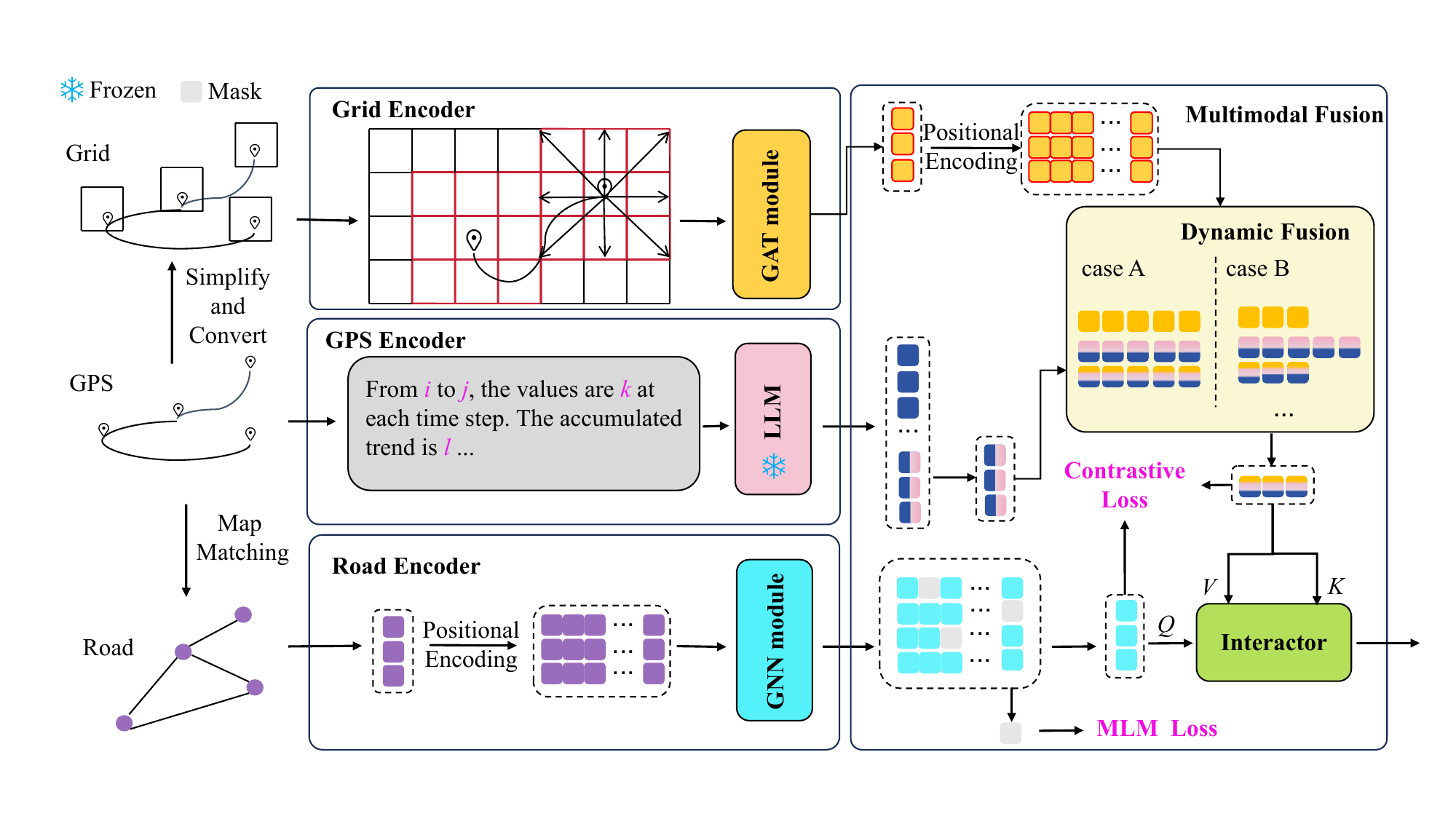}
  \caption{Overall architecture of the proposed MDTI framework.}
  \label{fig: model}
\end{figure*}

In this section, we present the proposed \name\ framework in detail. The overall architecture is shown in Figure~\ref{fig: model}.
To handle multimodal trajectory data, three dedicated encoders—Grid Encoder, GPS Encoder, and Road Encoder—are designed, providing the model with preliminary multimodal processing capability. Specifically, graph attention and graph neural networks are employed to encode map-matched road trajectories, while large-model semantic analysis enhances the representation of raw GPS data.
A multimodal fusion module is then introduced to align and integrate information across modalities, producing unified embeddings for TTE. In addition, we propose a dynamic trajectory alignment module that adaptively enriches the information density of embeddings, reducing redundancy and improving the accuracy of downstream TTE tasks. Finally, two complementary loss functions are applied to stabilize training and optimize performance. The following subsections provide a detailed description of each module in the \name\ framework.

% We now provide a detailed introduction of the proposed \name\ framework. 
% Figure~\ref{fig:model} presents the overall architecture, which mainly consists of three trajectory encoding modules, 
% two fusion modules, and two self-supervised learning tasks to optimize the training of \name.

\subsection{Grid Encoder}
\label{sec:grid_encoder}
% We design a Grid Encoder to learn spatio-temporal representations of grid trajectories. It converts the grid into a graph to model spatial adjacency, aggregates neighborhood information for node embeddings, and fuses them with temporal and trajectory features.

% The existing grid modeling approach only uses cell IDs to represent each grid and its internal trajectory points, which leads to information sparsity and fails to effectively capture the adjacency relationships between cells. To address this challenge, we introduce a graph attention mechanism (GAT)-based embedding learning method to learn dense embedding representations for grid cells. Unlike traditional static lookup tables, the embeddings of grid cells are dynamically computed by GAT based on input grid features, such as speed, time, and others. To further enhance the model's performance, we combine the 8-neighborhood connectivity rule with GAT, and design a grid trajectory modeling module to better capture the local relationships between grid cells .
Conventional grid-based modeling represents each cell and its internal trajectory points only by cell IDs, leading to sparse representations and limited adjacency awareness. To address this, we introduce a graph attention network (GAT)-based~\cite{velivckovic2017graph} embedding module that dynamically computes cell embeddings using input features such as speed and time. An 8-neighborhood connectivity scheme is adopted to better capture local spatial dependencies among grid cells.

Directly modeling a city-scale trajectory graph is computationally prohibitive. We therefore adopt a local subgraph strategy: for each trajectory point, we build a $3 \times 3$ grid-centered subgraph consisting of the current cell and its eight neighbors. This leverages the spatial locality of vehicle motion—single-step transitions rarely extend beyond adjacent cells—so the local neighborhood suffices to capture short-range dependencies while preserving trajectory continuity. Formally, for each node $i$, we retain edges only to itself and its eight immediate neighbors, yielding a first-order adjacency matrix  $A$:

\begin{equation}
A_{ij} =
\begin{cases}
1, & j=i \ \text{or}\ j \in \mathcal{N}(i), \\
0, & \text{otherwise}.
\end{cases}
\end{equation}

Here, $\mathcal{N}(i)$ denotes the 8-neighborhood of node $i$. This localized 
construction prevents the introduction of long-range, irrelevant edges and ensures that trajectory modeling relies primarily on adjacent regions.  

Given the graph structure \(\mathcal{G}_{\text{grid}} = (X, A)\), each grid cell is treated as a node, with its channel vector serving as the initial node feature. Based on the adjacency matrix
 $A$, nodes are connected to their direct neighbors to form the local subgraph.
 
% To capture spatial dependencies, neighborhood information is aggregated to update 
% node embeddings. For each node $i$, the embedding update is defined as:
To capture spatial dependencies, the node features are encoded using a two-layer GAT. The GAT iteratively aggregates information from neighboring nodes according to the attention mechanism. For the $l$-th GAT layer, the embedding of node $i$ is updated as:
\begin{equation}
h_i^{(l+1)} = \sigma \left( \sum_{j \in \mathcal{N}(i) \cup \{i\}} 
\alpha_{ij}^{(l)} W^{(l)} h_j^{(l)} \right),
\label{eq:gat_update}
\end{equation}
where $h_i^{(l)}$ is the embedding of node $i$ at layer $l$, $W^{(l)}$ is the learnable 
weight matrix, $\alpha_{ij}^{(l)}$ is the attention coefficient between nodes $i$ and $j$, 
and $\sigma(\cdot)$ is a nonlinear activation function. 

The attention coefficients are computed as:
\begin{equation}
\alpha_{ij}^{(l)} = \frac{\exp \left( \mathrm{LeakyReLU}\left( a^{\top} 
\left[ W^{(l)} h_i^{(l)} \,\Vert\, W^{(l)} h_j^{(l)} \right] \right) \right)}
{\sum_{k \in \mathcal{N}(i) \cup \{i\}} 
\exp \left( \mathrm{LeakyReLU}\left( a^{\top} 
\left[ W^{(l)} h_i^{(l)} \,\Vert\, W^{(l)} h_k^{(l)} \right] \right) \right)},
\label{eq:gat_attention}
\end{equation}
where $a$ is the learnable attention vector and $\Vert$ denotes vector concatenation.

% After aggregation, the embedding sequence is projected as:
% \begin{equation}
% G_{\text{emb}} = \text{Proj}\!\left( H^{(1)} \right) \in \mathbb{R}^{T \times D},
% \end{equation}
% where $H^{(1)}$ is the first-layer output, and $\text{Proj}(\cdot)$ denotes a linear projection. 
% where \(H^{(1)}\) is the first-layer output and \(\text{Proj}(\cdot)\) denotes a linear projection.  
% The resulting representation \( G_{\text{emb}} = \{ G_0, G_1, \ldots, G_T \} \) serves as the spatio-temporal embedding of grid nodes, where \(G_0\) denotes the [CLS] token and the remaining tokens correspond to different temporal segments of the trajectory mapped onto spatial grids.
After two successive GAT layers, the node embeddings are aggregated into
\begin{equation}
H^{(2)} = \sigma \!\left( 
    \text{GATConv}^{(2)} \!\left(
        \sigma \!\left( 
            \text{GATConv}^{(1)}(X, A)
        \right), A
    \right)
\right).
\end{equation}
Before applying the projection, we aggregate the node embeddings of all grid cells into the matrix form \(H\),
which serves as the grid-level embedding map for subsequent temporal modeling.
The embedding sequence is then projected as
\begin{equation}
G = \text{Proj}(H) \in \mathbb{R}^{T \times D},
\end{equation}
where \(\text{Proj}(\cdot)\) denotes a linear projection,
\(T\) is the number of temporal segments,
and \(D\) is the Transformer input embedding dimension.
% The resulting representation \( G_{\text{emb}} = \{ G_0, G_1, \dots, G_T \} \)
% serves as the spatio-temporal embedding sequence of grid nodes,
% with \(G_0\) corresponding to the [CLS] token, and the remaining tokens representing different time segments of the trajectory are mapped onto the spatial grids. This sequence embedding is then used as input for subsequent trajectory sequence modeling and Transformer-based encoding.

% To capture spatial dependencies, neighborhood information is aggregated to update node embeddings. For each 
% node $i$, the embedding update is formulated as:

% \begin{equation}
% h_i^{(1)} = \sigma \left( \sum_{j \in \mathcal{N}(i) \cup \{i\}} \alpha_{ij} W x_j \right),
% \end{equation}

% where $W$ is a linear transformation, $\alpha_{ij}$ denotes the attention weight, 
% and $\sigma(\cdot)$ is a nonlinear activation function. Stacking this operation 
% produces high-order node embeddings $H^{(1)}$, which are further projected to 
% the required dimension:

\subsection{GPS Encoder}

GPS data often exhibit spatiotemporal inconsistencies, including variations in timestamps, geographical locations, and sampling frequencies across trajectories. This heterogeneity poses major challenges for 
% travel time estimation
TTE, as conventional models struggle to capture inter-trajectory similarities and critical movement patterns from non-uniformly sampled data, resulting in reduced prediction accuracy. To address this issue, we explore the use of large language models (LLMs), leveraging their strong sequence modeling capabilities. However, applying LLMs directly to raw spatiotemporal trajectories faces a key obstacle: a modality gap between continuous coordinate values and the discrete token structure expected by LLMs, compounded by the absence of explicit semantic structure in raw GPS data. To bridge this gap, we propose a dedicated spatial discretization encoding method that maps continuous spatiotemporal coordinates into a sequence of discrete tokens.

% Formally, each trajectory is represented as a sequence of $T$ time steps with three numerical channels. 
Specifically, we decompose each trajectory into a sequence of structured textual segments with explicit semantics. Each trajectory is represented as a sequence of $T$ time steps, where each time step consists of three numerical channels, corresponding to the longitude, latitude, and temporal feature (e.g., timestamp or velocity). These channels jointly describe the spatiotemporal state of the moving object at each sampling point, serving as the fundamental input for local motion feature extraction. To extract local motion features, we divide the trajectory into non-overlapping chunks of length three, producing a set of sub-trajectories $\{ \mathcal{S}_m \}_{m=1}^{T/3}$. The choice of length three is based on empirical observations~\cite{rodrigues2020multi}: it is long enough to capture local motion patterns while remaining computationally efficient for large-scale datasets. Each sub-trajectory $\mathcal{S}_m$ is flattened into a 9-dimensional vector $\tilde{p}_m$. To capture representative spatial-temporal motion patterns, we compare each $\tilde{p}_m$ with a pattern library $P = \{P_k\}_{k=1}^{K} \in \mathbb{R}^{K \times 9}$, where $K$ denotes the number of predefined reference patterns. The similarity between $\tilde{p}_m$ and $P_k$ is computed as:
\begin{equation}
S_{m,k} = \exp \left( -\frac{\| \tilde{p}_m - P_k \|_2}{2 \sqrt{3}} \right).
\end{equation}
% The similarity scores are normalized with a softmax to yield probabilities $\pi_{p,:}$. For each segment, the top-3 prototypes and their scores are retained. In addition, we extract the raw value sequence $\{v_t\}$ and compute the cumulative trend
The formula converts the similarities between each sub-trajectory and the prototypes into comparable weights by exponentiating the Euclidean distances, thereby identifying the most representative motion patterns. To further enhance semantic expressiveness, we also extract cumulative trend quantities from the original sequence:

\begin{equation}
\Delta = \sum_{t=2}^{T} (v_t - v_{t-1}),
\label{eq:trend}
\end{equation}
% These features are then linearized into a structured natural language description, forming a textual prompt that integrates both numerical patterns and prototypical evidence. For example, a prompt may include the time indices, observed values, the cumulative trend, and identifiers of the top-3 matched prototypes.
% The prompts are tokenized and processed by a large pre-trained language model, which captures semantic dependencies across tokens and is able to extract latent relationships among different trajectory embeddings. 
% This metric characterizes the overall directional change of the sub-trajectory at the value level, compensating for dynamic information that may be overlooked by purely matching prototypes.
% On this basis, we integrate the timestep, observations, trend quantities, and the matched prototype identifiers into natural-language prompts. 
where $v_t$ represents the observed motion value at timestamp $t$.

After obtaining the structured motion information, we construct a natural-language prompt that integrates the temporal indices, cumulative trend, and pattern similarity results from Eq. \ref{eq:trend}. The prompt for the $i$-th trajectory segment is formulated as:
\begin{equation}
\text{Prompt}_i = f(\text{index}, \text{values}, \Delta, \text{patterns}, S_{i,k}),
\end{equation}
where $f(\cdot)$ denotes the textual formatting function that converts numerical features into natural-language sentences.
After feeding the prompts into a pre-trained language model, we take the embedding of the final token as the semantic representation of the current segment. Specifically, the embedding of the last token from the language model $\mathrm{LM}(\cdot)$ output serves as the representation vector for the segment:

\begin{equation}
z_i = \text{LastToken}\big(\mathrm{LM}(\text{Prompt})\big) \in \mathbb{R}^{d_{\text{LM}}}.
\label{eq:embedding}
\end{equation}

% By aggregating all entities in a batch, we obtain the GPS modality representation:
% \begin{equation}
% Z = \{ z_1, z_2, \ldots, z_N \},
% \end{equation}
% which captures the real motion patterns of the vehicle in continuous space.

This design combines prototype-based alignment, structured prompt generation, and representation learning with large-scale language models. It enables GPS trajectories to provide fine-grained motion semantics while leveraging the model’s capacity to capture semantic dependencies and extract latent relationships across trajectories, thereby enhancing the overall framework's expressive power.

\subsection{Dynamic Fusion}
Traditional cross-attention fusion mechanisms enable interaction between modalities but suffer from inherent asymmetry: one modality dominates as the query source while the other serves as the key-value counterpart, leading to imbalanced fusion and difficulty adapting to varying sequence lengths and structures. To address this, we propose a dynamic fusion mechanism that replaces the dominant–subordinate paradigm with symmetric and adaptive deep fusion, allowing bidirectional feature enhancement through fine-grained alignment and maximizing complementary information integration. In our framework, the trajectory representation generated by the GPS encoder acts as a prompt that guides and reinforces sequence learning in the grid encoder. Unlike conventional methods that simply concatenate or append GPS features to downstream predictors, our Dynamic Fusion module operates within the encoding stage, establishing early cross-modal correspondences between grid-based and GPS-based representations to ensure spatial and temporal consistency across modalities.

To guarantee that the two modalities can be effectively fused, we first apply a linear projection that maps GPS embeddings into the same latent space as the grid encoder output:
\begin{equation}
z'_t = \text{Proj}(z_t), \quad t = 1, \ldots, T.
\end{equation}

At the level of length, we introduce an alignment strategy to ensure consistent mapping between GPS sequences and grid tokens. When the GPS sequence is longer than the grid sequence, redundant temporal information is removed through truncation; when it is shorter, a zero-padding strategy is applied to preserve one-to-one correspondence. This mechanism achieves precise timestep synchronization across modalities and guarantees temporal consistency between GPS cues and grid tokens, providing a solid representation basis for subsequent fusion.

Finally, the aligned GPS sequence $z'$ is injected into the grid sequence via a residual fusion mechanism:
\begin{equation}
\tilde{G}_t = G_t + z'_t, \quad t = 1, \ldots, T,
\end{equation}
where $\tilde{G}_t$ denotes the fused representation. 
This design preserves the spatial modeling capacity of the grid encoder while explicitly incorporating GPS prompts, thereby improving the consistency and alignment of the learned representations.

\subsection{Road Encoder}

The road modality provides topological constraints from the underlying road network, which are essential for TRL. Unlike GPS points that directly capture continuous spatio-temporal observations, road sequences are naturally embedded in a graph structure, where adjacency and accessibility are governed by the topology. To effectively capture such structured dependencies, we adopt a hybrid design: a GAT on the graph side to update road node embeddings, and a Transformer encoder on the sequence side to model temporal dependencies within trajectories.

% On the graph side, each road node feature $h_i$ is first projected into a unified dimension through a linear layer, and then updated by stacked GATConv layers. For a given node $i$, its updated embedding is computed as:
% \begin{equation}
% h_i' = \sigma\!\left( \sum_{j \in \mathcal{N}(i)} \alpha_{ij} W h_j \right),
% \label{eq:gat_update}
% \end{equation}
% where $\mathcal{N}(i)$ denotes the set of neighbors of node $i$, $W$ is a learnable weight matrix, and $\sigma$ is a non-linear activation. The attention coefficient $\alpha_{ij}$ is defined as:
% \begin{equation}
% \alpha_{ij} = \frac{\exp\!\left(\text{LeakyReLU}\big(a^\top [Wh_i \,\|\, Wh_j]\big)\right)}{\sum_{k \in \mathcal{N}(i)} \exp\!\left(\text{LeakyReLU}\big(a^\top [Wh_i \,\|\, Wh_k]\big)\right)},
% \label{eq:gat_alpha}
% \end{equation}
% where $a$ is a learnable parameter vector and $\|\,$ denotes vector concatenation. Through this mechanism, each node representation incorporates both its own features and topologically weighted information from its neighbors, resulting in graph embeddings $g_{emb}$.
On the graph side, each road node feature $h_i$ is first projected into a unified 
embedding space through a linear transformation and then updated using stacked 
GATConv layers. The update process follows the same attention-based aggregation 
formulation as defined in Section~\ref{sec:grid_encoder} (Eq.~\ref{eq:gat_update}--\ref{eq:gat_attention}).  

Unlike the grid encoder, where the neighborhood $\mathcal{N}(i)$ is defined by fixed 
spatial proximity in a $3 \times 3$ grid, the road encoder constructs 
$\mathcal{N}(i)$ according to the topological connectivity of the road network. 
This allows message passing along physically connected road segments rather than 
adjacent spatial grids. Through this topology-aware aggregation, each node representation 
integrates its own attributes and the structural dependencies of neighboring roads, 
producing road-aware graph embeddings.

On the sequence side, we construct the road sequence embedding table by augmenting three special tokens \texttt{[PAD]}, \texttt{[CLS]}, and \texttt{[MASK]}. Given a trajectory, we obtain the corresponding sequence embeddings and further add temporal embeddings (week index and minute index), type embeddings, and positional encodings $\mathrm{PE}(\cdot)$ to preserve ordering. A padding mask ensures that attention is only applied to valid road segments.

The sequence encoder consists of stacked Transformer layers. In each layer, self-attention is first performed on the type embeddings, producing attention scores that are treated as adaptive priors $B$. These priors are then injected into the main self-attention computation:
\begin{equation}
\text{Attn}(Q,K,V) = \text{Softmax}\!\left(\frac{QK^\top}{\sqrt{d_k}} + B\right)V,
\label{eq:road_attn}
\end{equation}
which explicitly incorporates structural priors into the trajectory modeling process. After multiple layers, the output sequence representations are projected into a shared space, and the first \texttt{[CLS]} token is taken as the trajectory-level representation.
Finally, we denote the encoded road sequence representation as
\begin{equation}
    \mathbf{R} = \{\mathbf{R}_t\}_{t=1}^{T},
    \label{eq:road-output}
\end{equation}
where $\mathbf{R}_t$ represents the contextualized embedding of the $t$-th road segment after GAT-based graph encoding and Transformer modeling. 
The overall trajectory-level representation is derived from the \texttt{[CLS]} token of $\mathbf{R}$.

\subsection{Multimodal Interactor}

% A key challenge in spatiotemporal modeling is the effective fusion of heterogeneous modalities, such as GPS-based trajectory sequences and grid-based spatial features. Traditional fusion methods (e.g., cross-attention~\cite{vaswani2017attention}) often struggle to establish fine-grained spatiotemporal correspondences. This misalignment issue is particularly critical in travel time estimation. To address this, we design a modality interaction mechanism that explicitly models inter-modal dependencies using multi-head cross-attention, effectively resolving spatiotemporal misalignment caused by differences in sequence length and sampling frequency, and achieving fine-grained semantic alignment across modalities.
A central challenge in spatiotemporal modeling lies in the effective fusion of heterogeneous modalities, such as GPS-based trajectory sequences and grid-based spatial features. Traditional fusion approaches (e.g., cross-attention~\cite{vaswani2017attention}) often fail to establish fine-grained spatiotemporal correspondences, which is particularly detrimental for TTE. To overcome this, we design a modality interaction mechanism that leverages multi-head cross-attention to explicitly capture inter-modal dependencies, thereby mitigating sequence-length and sampling-frequency mismatches and achieving fine-grained semantic alignment across modalities.

Specifically, given a fused grid representation $\tilde{G} \in \mathbb{R}^{B \times L_g \times d}$ 
and a road sequence $\mathbf{R} \in \mathbb{R}^{B \times L_r \times d}$, we compute the cross-modal attention as:
\begin{equation}
\text{Attn}(\tilde{G}, \mathbf{R}) =
\text{Softmax}\left(
    \frac{(\tilde{G} W_Q)(\mathbf{R} W_K)^{\top}}{\sqrt{d_k}} + M_s
\right)(\mathbf{R} W_V),
\label{eq:cross_modal_attention}
\end{equation}
where \( W_Q, W_K, W_V \in \mathbb{R}^{d \times d_k} \) are learnable projection matrices, and \( M_s \) denotes the spatial alignment bias. 
In this design, the road sequence $\mathbf{R}$ acts as the \textit{query}, actively retrieving relevant contextual information, while the fused grid representation $\tilde{G}$ serves as the \textit{key} and \textit{value}, providing rich spatial semantics. 
This formulation enables adaptive and context-aware feature interaction between modalities, guided by the trajectory structure.

The fused output is then refined through a residual connection followed by a feed-forward network:
\begin{equation}
\hat{X} = \text{FFN}\left(\tilde{G} + \text{Attn}(\tilde{G}, \mathbf{R})\right),
\end{equation}
where \(\hat{X}\) denotes the final fused representation. 
This feed-forward refinement enhances nonlinear feature transformation and improves the overall trajectory modeling accuracy.
% Specifically, given a road sequence 
% $X \in \mathbb{R}^{B \times L_x \times d}$ 
% and a dynamically fused grid representation 
% $M \in \mathbb{R}^{B \times L_y \times d}$,
% we project them into query, key, and value spaces as follows:
% \begin{equation}
% Q = X W_Q, \quad K = M W_K, \quad V = M W_V,
% \end{equation}
% where 
% $W_Q, W_K, W_V \in \mathbb{R}^{d \times d_k}$ 
% are learnable parameter matrices. In this design, the road sequence serves as the query $Q$, actively retrieving relevant information, while the dynamically fused grid representation acts as the key $K$ and value $V$, providing a rich feature bank of spatial context. This arrangement enables the model to adaptively extract the most relevant spatiotemporal information from the grid features, guided by the road trajectory.

% The cross-modal interaction is computed via the attention mechanism:
% \begin{equation}
% \text{Attn}(X,M) = \text{Softmax}\left(\frac{QK^\top}{\sqrt{d_k}} + M_s \right) V,
% \end{equation}
% which resolves spatiotemporal misalignment through adaptive weight assignment. Residual connections and layer normalization further ensure training stability:
% \begin{equation}
% \hat{X} = \text{LayerNorm}(X + \text{Attn}(X,M)).
% \end{equation}

% The final fused representation $H^{fusion} = \hat{X}$ significantly improves the accuracy of trajectory modeling.

\subsection{Self-supervised Training}

\textbf{Contrastive Loss.} To learn effective cross-modal representations between prompt-enhanced grid embedding and road network modalities, we employ a contrastive framework~\cite{radford2021learning} that encourages alignment between semantically corresponding trajectory representations. Given a batch of trajectory pairs, where each sample contains both prompt-enhanced grid embedding visual representation and road network sequence representation, we aim to learn a shared embedding space where corresponding modalities are pulled together while non-corresponding pairs are pushed apart.
% For each trajectory sample in a batch, we obtain normalized embeddings from both modalities:

% \begin{equation}
% \mathbf{g}_i = \text{Normalize}(\text{GridEncoder}(\tilde{\mathbf{x}}_{\text{grid}}^{i})),
% \end{equation}

% \begin{equation}
% \mathbf{r}_i = \text{Normalize}(\text{RoadEncoder}(\mathbf{x}_{\text{road}}^{i})).
% \end{equation}

The contrastive learning objective employs InfoNCE loss with temperature scaling:

\begin{equation}
\mathcal{L}_{CL} = \frac{1}{2} \left[ \mathcal{L}_{\tilde{G} \rightarrow \mathbf{R}} + \mathcal{L}_{\mathbf{R} \rightarrow \tilde{G}} \right],
\end{equation}
where
\begin{align}
\mathcal{L}_{\tilde{G} \rightarrow \mathbf{R}} 
&= -\frac{1}{N} \sum_{i=1}^{N} 
\log \frac{\exp(\tilde{G}_i \cdot \mathbf{R}_i / \tau)}
{\sum_{j=1}^{N} \exp(\tilde{G}_i \cdot \mathbf{R}_j / \tau)}, \\
\mathcal{L}_{\mathbf{R} \rightarrow \tilde{G}} 
&= -\frac{1}{N} \sum_{i=1}^{N} 
\log \frac{\exp(\mathbf{R}_i \cdot \tilde{G}_i / \tau)}
{\sum_{j=1}^{N} \exp(\mathbf{R}_i \cdot \tilde{G}_j / \tau)}.
\end{align}

Here, $\tau$ is a learnable temperature parameter that controls the concentration of the distribution, and $N$ is the batch size. The bidirectional formulation ensures symmetric learning between both modalities.

\textbf{Masked Language Modeling Loss (MLM Loss).} 
To capture sequential dependencies and contextual relationships in road network trajectories, 
we adopt a BERT-style masked language modeling strategy~\cite{devlin2019bert}. 
In each trajectory sequence, 15\% of the road segments are randomly masked, 
and the model is trained to recover the original road IDs based on surrounding context 
and cross-modal information from the \textit{prompt-enhanced grid representations}. 
This design encourages the model to jointly leverage contextual and cross-modal cues 
while mitigating overfitting to the \texttt{[MASK]} token.  

For a masked road trajectory 
$\tilde{\mathbf{s}} = [s_1, s_2, \ldots, \texttt{[MASK]}, \ldots, s_L]$, the model predicts the original road segment IDs at the masked positions. 
The training objective is defined as:  
\begin{equation}
\mathcal{L}_{\text{MLM}} 
= -\frac{1}{|M|} \sum_{i \in M} 
\log P(s_i \,|\, \tilde{\mathbf{s}}, 
\mathbf{\tilde{G}, R )},
\label{eq:mlm}
\end{equation}
where $M$ denotes the set of masked positions, and 
$P(s_i \,|\, \tilde{\mathbf{s}}, 
\mathbf{\tilde{G}, R )}$ 
represents the probability of correctly predicting the road segment $s_i$ 
given the masked sequence $\tilde{\mathbf{s}}$ and its corresponding prompt-enhanced grid context.  

This objective enables the model to learn rich contextual representations 
of road networks while leveraging spatial semantics from the prompt-enhanced grid modality, 
facilitating a deeper understanding of trajectory patterns and spatial dependencies.

\textbf{Joint Training Objective.}
The final objective is the sum of the two self-supervised tasks:
\begin{equation}
\label{eq:total_loss}
\mathcal{L}_{\text{total}}=\mathcal{L}_{\text{CL}}+\mathcal{L}_{\text{MLM}}.
\end{equation}
Both losses contribute equally, enabling the model to learn cross-modal alignment and intra-modal sequential patterns simultaneously.

\section{Experiments}

\subsection{Experiment Settings}

We evaluate the performance of MDTI on three real-world datasets. The experiments are designed to address the following research questions:

\begin{itemize}[itemsep=0pt,parsep=0pt,topsep=0pt,leftmargin=*]
\item \textbf{RQ1}:
% Compared with various baseline methods, does MDTI significantly improve the performance of trajectory prediction?
Does MDTI provide more accurate estimates of travel time compared to various baseline methods?
\item \textbf{RQ2}: 
How do different components of the MDTI model contribute to the overall performance?
\item \textbf{RQ3}: 
% Under varying trajectory lengths, does the performance of MDTI highlight the importance of dynamic trajectory modeling?
Can the dynamic trajectory modeling module of MDTI adapt to inputs of different trajectory lengths?
\item \textbf{RQ4}: How is the transferability of MDTI?
\end{itemize}
Due to space limitations, we have placed RQ3 and RQ4 in the Appendix.

\textbf{Datasets.} 
We evaluate the proposed approach on three real-world multi-modal trajectory datasets collected from Porto, Chengdu, and Xi’an. Each dataset comprises GPS trajectories, route trajectories, and corresponding road networks. The GPS data are sourced from publicly available datasets released by Didi Chuxing\footnote{https://outreach.didichuxing.com/}, while the road networks are extracted using OSMNX~\cite{boeing2017osmnx}, which provides rich attributes such as road type, length, lane count, and topological connectivity.
For model training, we utilize only the topological structure of the road networks, differing from some baseline configurations that incorporate additional road attributes. The raw GPS trajectories are map-matched to the corresponding road networks through a standard map-matching algorithm, resulting in route trajectories and an assignment matrix. The assignment matrix explicitly encodes the mapping between trajectory sub-segments and their associated road segments.
For data partitioning, all datasets follow a consistent split across cities: 60\% for training, 20\% for validation, and 20\% for testing.

\textbf{Implementation Details.} Training is performed with a batch size of 32 for 30 epochs. We adopt Adam~\cite{kingma2014adam} as the optimizer with an initial learning rate of $2\times10^{-4}$, cosine annealing schedule, 10 warm-up epochs, a minimum learning rate of $1\times10^{-6}$, and weight decay of $1\times10^{-4}$. Both hidden and output embedding dimensions are fixed at 256 with a dropout rate of 0.1. For GPS trajectories, each sequence is segmented into patterns and compared with historical keys, and the most relevant ones are transformed into natural language prompts. These prompts are encoded by a pre-trained GPT-2 model, where the hidden state of the last token is taken as the trajectory embedding. Grid and road representations are generated by Transformer and GAT layers, while the cross-modal module aligns multi-source features. The objective combines multiple loss terms with weights of relation=0.5, step=1.0, rel=1.0, and cos=1.0.  

\textbf{Evaluation Metrics.}
We adopt two widely-used regression metrics: Mean Absolute Error (MAE) for measuring average prediction error magnitude, Mean Absolute Percentage Error (MAPE) for assessing relative error percentage. These metrics collectively provide a comprehensive assessment of prediction accuracy and model robustness.

\textbf{Baselines.}
We compare the proposed \name\ model with eight baseline methods, which can be categorized into four groups. All methods are trained using the same number of trajectories. The four groups include: one method based on GPS trajectories (e.g., Traj2vec~\cite{yao2017trajectory}); one method based on grid trajectories (e.g., TrajCL~\cite{chang2023}); four methods based on road trajectories (e.g., JCLRNT~\cite{mao2022jointly}, START~\cite{jiang2023self}, JGRM~\cite{ma2024more}, GREEN~\cite{zhou2025grid}); and two methods related to travel time estimation (e.g., DutyTTE~\cite{mao2025dutytte} and MulT-TTE~\cite{liao2024multi}). Further details about the baselines are provided in Appendix \ref{baseline}.

\subsection{Performance Comparison (RQ1)}
Table \ref{tab:performance_comparison} presents the overall performance of our model and baseline methods across the three city datasets. Several key observations emerge:
\textbf{(1) Our method consistently achieves the best MAE on all datasets and the lowest MAPE on Chengdu and Xi’an.} This confirms that the proposed cross-modal interaction module and modality-specific encoders within a unified feature space effectively enhance cross-domain alignment and semantic representation. In addition, the dynamic trajectory feature extraction mechanism adaptively balances information density across trajectories of varying lengths, mitigating local embedding bias.
\textbf{(2) Among single-modality trajectory modeling methods, Traj2vec--which directly encodes raw GPS point sequences--performs the worst, as RNN-based sequence modeling amplifies local noise.} In contrast, TrajCL, which relies solely on grid representations, suffers from spatial discretization errors and the loss of fine-grained positional information. Road-network-based methods generally outperform both, indicating that the topological connectivity of road segments and intersections provides strong structural constraints, while road semantics and traffic rules capture more stable spatiotemporal patterns.
\textbf{(3) On the Porto dataset, our method performs slightly worse than MulT-TTE in terms of MAPE, likely due to MAPE’s sensitivity to relative errors.} Nevertheless, our approach achieves the lowest MAE, demonstrating superior robustness to outliers and measurement noise. DutyTTE shows the weakest overall performance, as it lacks multimodal alignment and focuses primarily on global path consistency and interval coverage.
\textbf{(4) The cross-modal trajectory representation learning methods outperform single-modality ones, confirming that multimodal information effectively compensates for modality-specific limitations.} Our dynamic fusion mechanism adaptively regulates information density according to trajectory length, preventing feature over-concentration in local segments. This design yields consistently stable performance across both short and long trajectories while reducing the impact of extreme values.

\begin{table}[t]
\centering
\caption{Performance comparison across three datasets. The best results are in \textbf{bold} and the second-best are \underline{underlined}. $\downarrow$ means lower is better. Statistical significance was tested using paired t-tests across datasets; the differences were significant with $p < 0.05$.}
\label{tab:performance_comparison}
\adjustbox{max width=\linewidth}{
\begin{tabular}{@{}lcccccc@{}}
\toprule
& \multicolumn{2}{c}{Porto} & \multicolumn{2}{c}{Xi'an} & \multicolumn{2}{c}{Chengdu} \\
\cmidrule(lr){2-3}\cmidrule(lr){4-5}\cmidrule(lr){6-7}
Method & MAE$\downarrow$ & MAPE$\downarrow$ 
       & MAE$\downarrow$ & MAPE$\downarrow$ 
       & MAE$\downarrow$ & MAPE$\downarrow$ \\
\midrule
\multicolumn{7}{l}{\textbf{GPS-based trajectory representation learning}} \\
Traj2vec & 4.027 & 0.534 & 1.651 & 0.352 & 2.481 & 0.267 \\
\midrule
\multicolumn{7}{l}{\textbf{Grid-based trajectory representation learning}} \\
TrajCL   & 2.085 & 0.236 & 1.632 & 0.321 & 1.609 & 0.536 \\
\midrule
\multicolumn{7}{l}{\textbf{Travel Time Estimation Models}} \\
DutyTTE  & 3.780 & 0.303 & 4.131 & 0.302 & 4.525 & 0.316 \\
MulT-TTE & 1.524 & \textbf{0.147} & \underline{1.231} & \underline{0.172} & 1.461 & 0.234 \\
\midrule
\multicolumn{7}{l}{\textbf{Road-based trajectory representation learning}} \\
START    & 1.541 & 0.173 & 1.764 & 0.282 & 1.359 & 0.218 \\
JCLRNT   & 3.065 & 0.327 & 1.638 & 0.272 & 1.352 & 0.241 \\
JGRM     & 2.993 & 0.375 & 1.457 & 0.333 & 1.374 & 0.360 \\
GREEN    & \underline{1.510} & 0.165 & 1.241 & 0.184 & \underline{1.322} & \underline{0.185} \\
\midrule
MDTI     & \textbf{1.419} & \underline{0.152} & \textbf{1.169} & \textbf{0.170} & \textbf{1.143} & \textbf{0.183} \\
\bottomrule
\end{tabular}
}
\end{table}

\subsection{Component Contribution (RQ2)}
\textbf{Ablation Settings.}
To assess the contribution of each MDTI module, we conduct six ablations:
\begin{itemize}[itemsep=0pt,parsep=0pt,topsep=0pt,leftmargin=*]
\item \textbf{w/o GPS Enc.} Removes the GPS modality and the associated large language model, retaining only the road network and grid representations.
\item \textbf{w/o Road Enc.} Removes the road network encoder, leaving only the GPS and grid encoders.
\item \textbf{w/o Grid Enc.} Removes the grid modality; the embeddings generated by the GPS encoder are fused directly with the road network features.
\item \textbf{w/o Dynamic Fusion.} This variant removes the prompt-based Dynamic Fusion mechanism.
\item \textbf{w/o CL Loss.} Removes the contrastive loss.
\item \textbf{w/o MLM Loss.} Removes the MLM loss.
\end{itemize}
The outcomes of the ablation study conducted on the Porto dataset, as reported in Table~\ref{tab:ablation}, quantitatively validate the contribution of each component within the \textbf{MDTI} framework.
% Due to space limitations, the ablation results for MLM Loss and CL Loss are presented in Appendix \ref{sec:appendix_ablation}.

\textbf{Importance of Multimodal Encoders.}
Removing any single modality encoder results in a consistent performance decline across all evaluation metrics (MAE and MAPE), highlighting the necessity of multimodal information fusion. Notably, the exclusion of the Grid Encoder yields the largest performance drop, indicating that grid-based spatial representations provide crucial contextual cues that complement the sequential dynamics captured by GPS trajectories and the topological structure encoded by the road network. The observed degradation when removing either the GPS Encoder or the Road Encoder further confirms that both trajectory dynamics and road topology contribute distinct and indispensable information for accurate TTE.

\textbf{Effectiveness of Pre-training Objectives.}
Ablating the two self-supervised pre-training tasks also confirms their essential roles. Removing the MLM loss leads to a more pronounced degradation than removing the CL loss, suggesting that the MLM task—which reconstructs masked road segments based on contextual and grid information—plays a key role in modeling intra-modal dependencies and cross-modal interactions. The relatively smaller impact of removing the CL loss indicates that while inter-modal contrastive alignment enhances representation consistency, the model can still acquire meaningful embeddings through the MLM objective and supervised training. Nevertheless, the best results are achieved when both objectives are jointly optimized, demonstrating their complementary effects in improving representation quality and downstream performance.

\textbf{Indispensability of the Dynamic Fusion Alignment Mechanism.}
The experimental results clearly demonstrate that the proposed dynamic fusion alignment mechanism is crucial for maintaining the overall performance of the model. Its removal leads to a substantial drop across all evaluation metrics, producing outputs markedly inferior to those of the full model that benefits from dynamic feature alignment. Although some global metrics may appear comparable, qualitative inspection reveals evident spatial distortions and implausible artifacts in the predicted distributions. These findings confirm that the primary contribution of the dynamic fusion alignment mechanism lies in its ability to intelligently align and integrate multi-source features, ensuring spatial coherence and physical plausibility in model predictions, rather than merely optimizing global numerical accuracy.

\begin{table}[h]
\centering
\caption{Ablation study of different modules.}
\begin{tabular}{lccc}
\midrule
\textbf{} & MAE$\downarrow$  & MAPE$\downarrow$ \\
\midrule
w/o GPS enc. & 1.441  & 0.153 \\
w/o Road enc. & 1.468  & 0.156 \\
w/o Grid enc. & 1.593 & 0.180 \\
w/o CL loss & 1.513 & 0.165  \\
w/o MLM loss & 1.540 & 0.167 \\
w/o Dynamic FusionAlign & 1.457 & 0.154 \\
\midrule
\textbf{MDTI} & \textbf{1.419} & \textbf{0.152} \\
\midrule
\end{tabular}
\label{tab:ablation}
\end{table}

\section{Related Work}
\subsection{Multimodal Trajectory Representation Learning}
TRL encodes sequences of states, actions, or spatiotemporal points into compact vectors that preserve temporal structures and semantic information. Early work introduced Traj2vec~\cite{yao2017trajectory}, which directly modeled raw GPS trajectories as point sequences to derive compact representations for downstream tasks. However, due to inherent issues in GPS data—such as positioning drift, U-turns, and redundant sampling—these approaches often produced unstable and poorly transferable representations~\cite{li2008mining}.

To overcome these limitations, grid-based trajectory modeling methods have been proposed~\cite{chang2023,li2018deep}. By discretizing the spatial domain into grid cells and constructing graph-based structures, these approaches enhance the capacity for semantic and structural trajectory modeling. For instance, TrajCL~\cite{chang2023} adopts contrastive learning to model spatial correlations and structural relationships among trajectories jointly.

In parallel, road-based trajectory modeling has emerged as another major direction. Using map-matching techniques~\cite{lou2009map}, trajectories are aligned with road networks, which introduces topological constraints and enables richer representations of travel patterns. JCLRNT~\cite{mao2022jointly} generates contrastive samples between roads by leveraging contextual neighbors and constructing trajectory-level contrasts through detour path replacement. START~\cite{jiang2023self} incorporates temporal regularities and travel semantics into a self-supervised framework, improving both representation quality and generalization. JGRM~\cite{ma2024more} further strengthens spatiotemporal road features by mapping raw GPS points to corresponding road segments\cite{kanoulas2006finding,yuan2012discovering,zhu2025learning}. Based on this trend, GREEN~\cite{zhou2025grid} integrates grid-based and road-based representations to exploit their complementarity. 

Building on these works, we unify GPS-based, grid-based, and road-based information in a multimodal framework to mitigate single-modality limitations for TTE.

\subsection{Travel Time Estimation}
Travel time estimation refers to predicting the time required to traverse a route or road segment from origin to destination. Existing studies can be broadly categorized into path-based methods and origin–destination (OD)-based approaches.

Path-based methods~\cite{ta2022adaptive, li2023physical} use complete trajectories or trajectory segments as input, enabling the modeling of vehicle dynamics within the road network. These approaches typically leverage road segment features, trajectory sequences, or path structures to capture spatiotemporal dependencies. For instance, DeepTTE~\cite{wang2018will} integrates graph convolutional networks with LSTM~\cite{yao2017deepsense} to model the spatiotemporal dependencies of GPS sequences. HierETA~\cite{chen2022interpreting} introduces a multi-view hierarchical choice model for fine-grained and interpretable TTE. MulT-TTE~\cite{liao2024multi} jointly models trajectory, attribute, and semantic sequences, and further applies self-supervised learning to improve generalization, yielding significant gains in prediction accuracy.

OD-based methods~\cite{mao2025dutytte, liu2025spottrip}, by contrast, rely only on OD pairs and departure times, avoiding the need for full trajectory inputs. More recently, DutyTTE~\cite{mao2025dutytte} combines reinforcement learning with a mixture-of-experts framework to optimize path prediction and capture segment-level uncertainty, enabling confidence interval estimation.

Our work is more closely aligned with path-based approaches, as it exploits the spatiotemporal information embedded in complete trajectories to achieve more accurate and robust TTE.
\section{Conclusion}
This paper proposes a multimodal dynamic trajectory representation framework, \textbf{MDTI}, which integrates GPS sequences, grid trajectories, and road network constraints to learn diverse trajectory representations. Corresponding encoders are designed for these three types of trajectories. In addition, to address the limitations of fixed-length feature vectors, a dynamic fusion module is developed. Finally, two loss functions are designed to enhance the framework’s performance. Experiments on three datasets demonstrate that \name\ outperforms state-of-the-art baseline methods.

\section{Acknowledgment}
This work was supported in part by the Zhejiang Province Basic Public Welfare Research Project under Grant LTGG24F020009, in part by the National Natural Science Foundation of China under Grant 62476247, in part by the "Pioneer" and "Leading Goose" R\&D Program of Zhejiang under Grant 2024C01214, in part by the Zhejiang Provincial Natural Science Foundation under Grant LR21F020003, and in part by the Zhejiang Provincial Natural Science Foundation of China under Grant No.LQN26F030027.

\newpage
\bibliographystyle{ACM-Reference-Format}
\bibliography{9.References}

% \newpage

\balance
\appendix
\section{Appendix}

\subsection{Notation}
We summarize and list all notations used throughout this paper in the following Table ~\ref{notation}.
\begin{table}[H]
\centering
\caption{Notation definitions used in this paper.}
\renewcommand{\arraystretch}{1.2}
\adjustbox{max width=\linewidth}{
\begin{tabular}{c >{\centering\arraybackslash}p{7cm}}

\toprule
\textbf{Notation} & \textbf{Description} \\
\midrule
\(\mathcal{T}^{\text{gps}}\)   & an ordered sequence of raw GPS points \\
\(\mathcal{T}^{\text{grid}}\)  & a grid trajectory \\
\(\mathcal{G}\)                & a road network \\
\(\mathcal{V}\)                & the set of vertices \\
\(\mathcal{E}\)                & the set of directed edges \\
\(\mathcal{T}^{\text{road}}\)  & a road trajectory \\
\(A\)                          & a first-order adjacency matrix \\
\(T\)                          & the number of temporal segments \\
\(S_m\)                        & a set of sub-trajectories \\
\(D\)                          & the embedding dimension of the Transformer input \\
\(\bar{K}\)                    & the number of predefined reference patterns \\
\(z'\)                         & projected GPS embedding sequence \\
\(\mathbf{R}\)                 & a road sequence \\
\(G\)                          & a grid embedding sequence \\
\(\tilde{G}\)                  & a fused grid representation \\
\bottomrule
\label{notation}

\end{tabular}
}
\end{table}

\subsection{Baseline}
\label{baseline}
The baseline methods employed in this paper are detailed as follows:
\noindent\textbf{1. GPS-based trajectory representation learning.}
\begin{itemize}[leftmargin=*]
\item \textbf{Traj2vec~\cite{yao2017trajectory}:} An RNN-based seq2seq model converts GPS trajectory into a feature sequence to learn trajectory representations.
\end{itemize}

\noindent\textbf{2. Grid-based trajectory representation learning.}
\begin{itemize}[leftmargin=*]
\item \textbf{TrajCL~\cite{chang2023}:} The state-of-the-art grid-based method proposes a set of trajectory augmentations on grid trajectory in free space and dual-feature self-attention to learn grid trajectory representations using contrastive learning with the Transformer.
\end{itemize}

\noindent\textbf{3. Road-based trajectory representation learning.}
\begin{itemize}[leftmargin=*]
\item \textbf{JCLRNT~\cite{mao2022jointly}:} A jointly contrastive learning framework that performs within-road, within-trajectory, and cross-scale road--trajectory contrast, maximizing mutual information to obtain robust representations.
\item \textbf{START~\cite{jiang2023self}:} A self-supervised trajectory representation model that incorporates temporal regularities and travel semantics using a graph attention network and a time-aware trajectory encoder, trained with span-masked recovery and contrastive learning tasks.
\item \textbf{JGRM~\cite{ma2024more}:} A joint GPS--route modeling approach with dual encoders and a cross-modal interaction module, trained in a self-supervised manner via masked language modeling (MLM) and cross-modal matching (CMM) objectives.
\item \textbf{GREEN~\cite{zhou2025grid}:} A multimodal trajectory representation method that jointly leverages grid-based and road-based trajectories, aligned and fused via contrastive learning and masked reconstruction objectives.
\end{itemize}

\noindent\textbf{4. Travel Time Estimation Models.}
\begin{itemize}[leftmargin=*]
\item \textbf{DutyTTE~\cite{mao2025dutytte}:} A reinforcement learning-based framework that enhances the alignment between predicted and actual travel routes, and employs a mixture-of-experts mechanism to quantify uncertainty across different road segments.
\item \textbf{MulT-TTE~\cite{liao2024multi}:} A multi-faceted model that decomposes routes into trajectory, attribute, and semantic sequences, combines sequential learning with Transformer encoders, and incorporates self-supervised masking and multi-task optimization.
\end{itemize}

% \begin{table}[H]
% \centering
% \caption{Ablation study of loss components.}
% \begin{tabular}{lccc}
% \midrule
% \textbf{} & MAE$\downarrow$  & MAPE$\downarrow$ \\
% \midrule
% w/o CL loss & 1.418  & \textbf{0.151} \\
% w/o MLM loss & 1.540 & 0.167 \\
% \midrule
% \textbf{MDTI} & \textbf{1.419} & 0.152 \\
% \midrule
% \end{tabular}
% \label{tab: additional ablation}
% \end{table}

\begin{table}[htbp!]
\centering
\caption{The transferability accuracy of MDTI across three cities.}
\begin{tabular}{lccc}
\midrule
\textbf{Scenario} & \textbf{MAE}$\downarrow$ & \textbf{RMSE}$\downarrow$ & \textbf{MAPE} $\downarrow$\\
\midrule
Porto $\rightarrow$ Chengdu & & & \\
\quad w/ Transfer & 1.157 & 1.634 & 0.193 \\
\quad w/o Transfer & 1.143 & 1.508 & 0.183 \\
\midrule
Chengdu $\rightarrow$ Xi'an & & & \\
\quad w/ Transfer & 1.153 & 1.568 & 0.173 \\
\quad w/o Transfer & 1.169 & 1.614 & 0.170 \\
\midrule
\end{tabular}
\label{tab:transfer_results}
\end{table}

% \subsection{Additional Ablation Studies}
% \label{sec:appendix_ablation}
% In Table~\ref{tab: additional ablation}, we further present two additional ablation experiments to examine the effects of removing the MLM loss and the CL loss, respectively.

% Ablating the two self-supervised loss components further validates their essential roles in the model. Removing the MLM loss results in a more substantial performance drop than removing the CL loss, indicating that the MLM objective—which reconstructs masked road segments using contextual and grid information—plays a crucial role in capturing intra-modal dependencies and enhancing spatial–temporal representations. In contrast, the smaller degradation observed when excluding the CL loss suggests that, although cross-modal alignment is weakened, the model can still learn meaningful embeddings through the MLM objective and supervised optimization. Nevertheless, the best performance is achieved when both loss functions are jointly optimized, confirming their complementary contributions to improving representation quality and downstream travel time estimation accuracy.

\subsection{Trajectory Length Impact (RQ3)}
\begin{figure}[htbp!]
    \centering
    \includegraphics[width=1\linewidth]{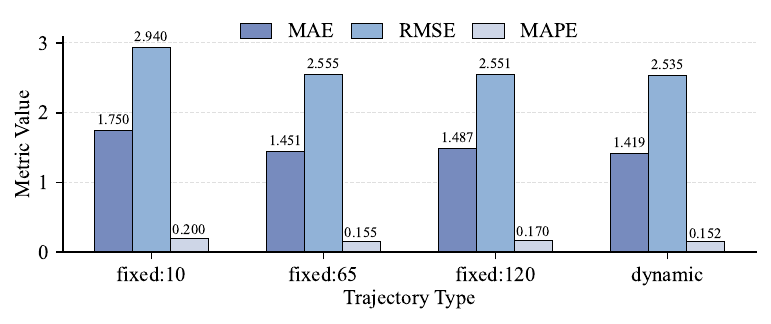}
    \caption{Comparison of fixed and dynamic trajectory lengths on prediction errors.}
    \label{fig: Length Impact}
\end{figure}

We analyze the effect of fixed and dynamic trajectory lengths on model performance by comparing prediction error metrics. Specifically, we evaluate three fixed lengths (10, 65, and 120) and one dynamic adjustment strategy, using MAE, RMSE, and MAPE as evaluation metrics. The results are presented in Figure~\ref {fig: Length Impact}.   

Under fixed-length conditions, prediction errors decrease as trajectory length increases. This phenomenon indicates that longer trajectories contain a more extensive history of object motion, thereby providing the model with more comprehensive information on motion patterns. In contrast, shorter trajectories provide limited information, making it difficult for the model to distinguish whether an object is moving at a constant speed, accelerating, or turning, which increases the uncertainty of prediction. Furthermore, longer trajectories demonstrate stronger noise resistance. Short trajectories are highly sensitive to measurement noise or transient irregular movements (e.g., obstacle avoidance), where even a single abnormal point can significantly disturb the judgment of the overall trend; meanwhile, longer trajectories, by virtue of incorporating more data points, can effectively smooth out random noise and allow the model to focus on stable and intrinsic motion trends.  

More importantly, the dynamic trajectory strategy consistently outperforms all fixed-length configurations across all error metrics. This result demonstrates that adaptive adjustment of the trajectory length allows the model to provide the most appropriate amount of contextual information for different motion scenarios, yielding optimal or near-optimal prediction performance. 

\subsection{Transferability Study (RQ4)}
\label{sec:appendix_4}
In Table \ref{tab:transfer_results}, we evaluate the transferability of our pre-trained model across distinct urban scenarios through a series of experiments. In particular, we fine-tuned the model on the TTE task to adapt it to different city environments, while employing the transferred prompt files during task training.  

The results demonstrate that transferring from Porto to Chengdu did not yield additional improvements; instead, the performance slightly declined. Nevertheless, the outcomes remain competitive compared with other baselines. This phenomenon may be attributed to the substantial differences in traffic patterns between the two cities, which limit the effectiveness of transfer learning. Conversely, transferring from Chengdu to Xi’an led to overall performance gains, suggesting that the two cities share more similar traffic characteristics, thereby enabling transfer learning to exert a positive effect.

\subsection{Implementation Details}
As a supplement to the implementation details in the main paper, we provide a detailed explanation of Algorithm~\ref{alg:pretrain}, which describes the multimodal pretraining process of MDTI.

In the initialization stage, we set the model parameters $\Theta$ (line 1) and prepare the road network graph $\mathcal{G}_{\text{road}}$, the grid graph $\mathcal{G}_{\text{grid}} = (X, A)$, and the GPS trajectories $\{\tau_i\}$. Next, in Stage~1 , for each training iteration we sample a mini-batch of trajectories and, for each trajectory $\tau_i$ (lines 2-3), construct its road segment sequence on $\mathcal{G}_{\text{road}}$. We then update the road node features with a GAT module and feed the resulting road sequence into a Transformer to obtain the road sequence representation $R_i$ (lines 4-8). Then, in Stage 2 , for each trajectory we extract local grid subgraphs along its path on $\mathcal{G}_{\text{grid}}$, apply a two-layer GAT on $(X, A)$ to obtain grid node embeddings $H$, and project them to form the grid sequence embeddings $G_i$ (lines 9-13). In Stage~3, we further process each GPS trajectory $\tau_i$ by segmenting it into sub-trajectories and computing motion patterns. These patterns are organized into structured prompts $\mathrm{Prompt}_{i,t}$, which are fed into a language model $\mathrm{LM}(\cdot)$ to obtain GPS token embeddings $Z_i = \{z_{i,t}\}_{t=1}^{T}$ (lines 14-18). Next, in Stage 4, we align $G_i$ and $Z_i$ into the same latent space and use a residual fusion module to obtain a fused grid representation $\tilde{G}_i$. We then apply cross-attention between the road sequence representation $R_i$ and $\tilde{G}_i$ to produce the fused trajectory representation $\hat{R}_i$ (lines 19-23). In Stage 5, we compute a contrastive loss $\mathcal{L}_{\mathrm{CL}}$ between the fused grid representations and the road sequence representations (line 24), encouraging trajectories from the same instance to be close while pushing apart those from different instances. Finally, in Stage 6 , we randomly mask a subset of road tokens in $\{R_i\}$, predict the masked tokens with the road encoder, and compute the MLM loss $\mathcal{L}_{\mathrm{MLM}}$. The total loss
$\mathcal{L} = \mathcal{L}_{\mathrm{CL}} + \mathcal{L}_{\mathrm{MLM}}$
is then used to update the model parameters $\Theta$ via gradient descent (lines 25-28), and this process is repeated until convergence.

\subsection{Case Study}
\begin{figure}[htbp]
    \centering
    \includegraphics[width=0.65\linewidth]{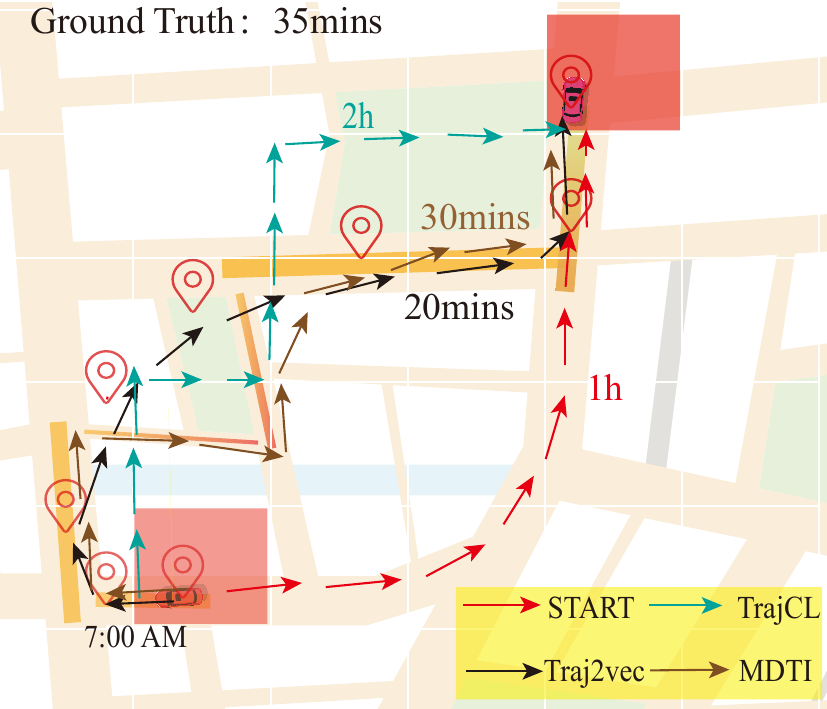}
    \caption{A case study of travel time estimation.}
    \label{fig: case study}
\end{figure}

We randomly selected ride-hailing trips in Porto as our testbed. We simulated travel-time estimation using different approaches and compared the estimated durations with the ground-truth travel times. As shown in Figure \ref{fig: case study}, we visualize a local region of Porto using three types of base maps: grids, GPS traces, and the road network. We use color-coded arrows to indicate the predictions produced by different models.

Most baselines tend to follow routes dictated by their input trajectory representation when estimating travel time. For example, the GPS-based Traj2vec plans routes according to GPS points, whereas the grid-based TrajCL infers routes based on grid transitions. In contrast, our method jointly leverages multiple trajectory modalities to identify the most plausible route, and its travel-time estimates are the closest to the ground truth. These results demonstrate the effectiveness of the proposed approach.

\begin{algorithm}[H]
\caption{Multimodal Pretraining Process}
\label{alg:pretrain}
\begin{algorithmic}[1]
    \REQUIRE 
        Road network graph $\mathcal{G}_{\text{road}}$, grid graph $\mathcal{G}_{\text{grid}} = (X, A)$, 
        GPS trajectories $\{\tau_i\}$, prototype pattern set $\mathcal{P}$, 
        batch size $N_b$, learning rate $\eta$.
    \ENSURE all trained model parameters $\boldsymbol{\Theta}$. 
    
    \STATE Initialize parameters $\boldsymbol{\Theta}$
    
    \FOR{Each training iteration}
    \STATE Sample a mini-batch of trajectories $\{\tau_i\}_{i=1}^{N_b}$
    
    \textbf{Stage 1: Road Trajectory Encoding:} 
   
    \FOR{Each trajectory $\tau_i$}
    \STATE Construct road segment sequence $\boldsymbol{r}_i$ from $\tau_i$ on $\mathcal{G}_{\text{road}}$
    \STATE Update road node features by GAT on $\mathcal{G}_{\text{road}}$
    \STATE Encode $\boldsymbol{r}_i$ with Transformer to obtain road sequence representation $\boldsymbol{R}_i$
    \ENDFOR

    \textbf{Stage 2: Grid Trajectory Encoding:} 

    \FOR{each trajectory $\tau_i$}
    \STATE Extract local grid subgraphs along $\tau_i$ and obtain grid node features $X$
    \STATE Apply a two-layer GAT on $(X, A)$ to compute grid node embeddings $\boldsymbol{H}$
    \STATE Project $\boldsymbol{H}$ into grid sequence embeddings $\boldsymbol{G}_i$
    \ENDFOR

    \textbf{Stage 3: GPS Trajectory Encoding:} 

    \FOR{Each trajectory $\tau_i$}
        \STATE Segment $\tau_i$ into sub-trajectories and compute motion patterns
        \STATE Form structured prompts $\text{Prompt}_{i,t}$ and feed them into language model $\mathrm{LM}(\cdot)$
        \STATE Obtain GPS token embeddings $\boldsymbol{Z}_i = \{\boldsymbol{z}_{i,t}\}_{t=1}^{T}$
    \ENDFOR

    \textbf{Stage 4: Embedding Fusion:} 

    \FOR{Each trajectory $\tau_i$}
        \STATE Align $\boldsymbol{G}_i$ and $\boldsymbol{Z}_i$ into the same latent space
        \STATE Fuse them via residual fusion to get fused grid representation $\tilde{\boldsymbol{G}}_i$
        \STATE Apply cross-attention between road sequence $\boldsymbol{R}_i$ and $\tilde{\boldsymbol{G}}_i$ to obtain fused trajectory representation $\hat{\boldsymbol{R}}_i$
    \ENDFOR

    \textbf{Stage 5: Contrastive:} 

    \STATE Compute contrastive loss $\mathcal{L}_{\text{CL}}$ between $\{\tilde{\boldsymbol{G}}_i\}$ and $\{\boldsymbol{R}_i\}$

    \textbf{Stage 6: Masked Language Modeling:} 

    \STATE Randomly mask a subset of road tokens in $\{\boldsymbol{R}_i\}$
    \STATE Predict the masked tokens with the road encoder and compute MLM loss $\mathcal{L}_{\text{MLM}}$
    \STATE Compute total loss $\mathcal{L} = \mathcal{L}_{\text{CL}} + \mathcal{L}_{\text{MLM}}$
    \STATE Update parameters $\boldsymbol{\Theta} \leftarrow \boldsymbol{\Theta} - \eta \nabla_{\boldsymbol{\Theta}} \mathcal{L}$
  
    \ENDFOR
    \STATE \textbf{return} all trained parameters $\boldsymbol{\Theta}$
\end{algorithmic}

\end{algorithm}

\clearpage

% To further analyze the model’s performance at the single-trip level, we randomly select two trajectories from the test set as case studies.

% \textbf{Case 1.}  
% The selected trajectory consists of 29 road segments, and the ground-truth travel time is 4.25 minutes.  
% As a simple baseline, we estimate the travel time by multiplying the number of segments by the average per-segment travel time computed from the training set. This baseline predicts 6.34 minutes, resulting in an absolute error of 2.09 minutes.  
% In contrast, our MDTI model predicts 4.83 minutes for the same trip, with an absolute error of only 0.58 minutes.  
% This result demonstrates that the pretrained trajectory and road-network representations enable the model to better capture local road conditions and spatiotemporal dynamics, leading to more accurate travel time estimation.

% \textbf{Case 2.}  
% Another randomly selected trajectory contains 36 road segments, and the ground-truth travel time is 5.75 minutes.  
% The simple baseline predicts 7.88 minutes, resulting in an absolute error of 2.13 minutes.  
% By comparison, the MDTI model predicts 5.41 minutes, with an absolute error of only 0.34 minutes.  
% This case further confirms that the pretrained MDTI framework effectively captures the underlying road conditions and traffic patterns, thereby improving the robustness and accuracy of travel time estimation at the single-trip level.

\end{document}